\documentclass[acmsmall,screen]{acmart}
\AtBeginDocument{%
  \providecommand\BibTeX{{%
    \normalfont B\kern-0.5em{\scshape i\kern-0.25em b}\kern-0.8em\TeX}}}

\begin{document}

\title{Redefining the Shortest Path Problem Formulation of the Linear Non-Gaussian Acyclic Model: Pairwise Likelihood Ratios, Prior Knowledge, and Path Enumeration}

\author{Hans Jarett J. Ong}
\email{ong.hans_jarett.ol5@naist.ac.jp}
\affiliation{%
  \institution{Nara Institute of Science and Technology}
  \city{Nara}
  \country{Japan}
  \postcode{630-0192}
}

\author{Brian Godwin S. Lim}
\email{lim.brian_godwin_sy.la6@naist.ac.jp}
\affiliation{%
  \institution{Nara Institute of Science and Technology}
  \city{Nara}
  \country{Japan}
  \postcode{630-0192}
}

\author{Renzo Roel P. Tan}
\email{rr.tan@is.naist.jp}
\affiliation{%
  \institution{Nara Institute of Science and Technology}
  \city{Nara}
  \country{Japan}
  \postcode{630-0192}
}
\affiliation{%
  \institution{Ateneo de Manila University}
  \city{Metropolitan Manila}
  \country{Philippines}
  \postcode{1108}
}

\author{Kazushi Ikeda}
\email{kazushi@is.naist.jp}
\affiliation{%
  \institution{Nara Institute of Science and Technology}
  \city{Nara}
  \country{Japan}
  \postcode{630-0192}
}


\begin{abstract}
Effective causal discovery is essential for learning the causal graph from observational data. The linear non-Gaussian acyclic model (LiNGAM), a causal discovery technique, operates under the assumption of a linear data generating process with non-Gaussian noise in determining the causal graph. Its assumption of unmeasured confounders being absent, however, poses practical limitations. In response, empirical research has shown that the reformulation of LiNGAM as a shortest path problem (LiNGAM-SPP) addresses this limitation. Within LiNGAM-SPP, mutual information is chosen to serve as the measure of independence. A challenge is introduced -- parameter tuning is now needed due to its reliance on $k$-nearest neighbors ($k$NN) mutual information estimators. The paper proposes a threefold enhancement to the LiNGAM-SPP framework. 

First, the need for parameter tuning is eliminated by using the pairwise likelihood ratio in lieu of $k$NN-based mutual information. This substitution is validated on a general data generating process and benchmark real-world datasets, outperforming existing methods especially when given a larger set of features. The incorporation of prior knowledge is then enabled by a node-skipping strategy implemented on the graph representation of all causal orderings to eliminate violations based on the provided input of relative orderings. Flexibility relative to existing approaches is achieved. Last among the three enhancements is the utilization of the distribution of paths in the graph representation of all causal orderings. From this, crucial properties of the true causal graph such as the presence of unmeasured confounders and sparsity may be inferred. To some extent, the expected performance of the causal discovery algorithm may be predicted. The aforementioned refinements advance the practicality and performance of LiNGAM-SPP, showcasing the potential of graph-search-based methodologies in advancing causal discovery.
\end{abstract}




\maketitle


\section{Introduction}
Machine learning has achieved significant success by its ability to identify patterns in large sets of independent and identically distributed data. However, many of these techniques face difficulty with out-of-distribution generalization because they tend to disregard important information such as interventions in the world, domain shifts, and temporal structure \cite{scholkopf_toward_2021}. Causal models, particularly graphical causal models \cite{pearl_causality_2000,pearl_causal_2016,peters_elements_2017}, offer a solution to this problem by modeling the data generating process itself using causal graphs. This enables causal models to generalize out of distribution and to make interventional and counterfactual predictions. Aside from graphical causality, another well-established framework for causal inference is the potential outcomes framework \cite{rubin_causal_2005, yao2021_survey_ci}. This paper, nonetheless, focuses solely on graphical causality.

While there are methods for estimating interventions and counterfactuals such as Pearl's do-calculus and structural causal models (SCMs) \cite{pearl_causality_2000, pearl_causal_2016, pearl_book_2018}, these require knowing the causal graph which is often unknown in practice. A straightforward approach to obtaining the causal graph is to construct it from domain knowledge. This process is often tedious and time-consuming, especially in settings with many features. Causal discovery methods resolve this by attempting to learn the causal structure from data. One such causal discovery method is the Linear Non-Gaussian Acyclic Model (LiNGAM) \cite{shimizu_linear_2006,shimizu_directlingam_2011, shimizu_lingam_2014}, which assumes a linear data generating process and noise terms with non-Gaussian distributions.

LiNGAM also assumes the absence of unmeasured confounders \cite{shimizu_linear_2006}, which can be a significant limitation in practice. Several improvements to LiNGAM have been proposed in order to address this issue, one of which is the formulation of LiNGAM as a shortest path problem (LiNGAM-SPP) \cite{suzuki_lingam_mmi, suzuki2024generalization}. LiNGAM-SPP identifies the causal ordering of the features by selecting the path that minimizes the mutual information of the noise terms. Although LiNGAM-SPP presents a promising solution to the challenge of unmeasured confounders, a notable drawback is its reliance on parameter tuning. This proves impractical in real-world scenarios where access to the underlying data generating process is unavailable. 

In light of this, we introduce an enhanced version of LiNGAM-SPP that eliminates the need for parameter tuning and at the same time demonstrates superior performance and computational efficiency. In addition to its primary objective of addressing unmeasured confounders, the LiNGAM-SPP offers a unique perspective on the causal ordering problem. It allows us to view this problem as a path search problem, where each path corresponds to a potential causal ordering. This paper also explores other ways to leverage this path formulation beyond its initial purpose. 

While the original LiNGAM-SPP papers \cite{suzuki_lingam_mmi, suzuki2024generalization} have established that each possible path corresponds to a causal ordering and that the shortest path corresponds to the most likely true causal ordering, we ask: ``Can we exploit other features of this path search space?'' The answer is affirmative. We show that constrained searches where certain nodes are excluded correspond to incorporating prior knowledge and that this method of integrating knowledge is more flexible than those employed by LiNGAM \cite{shimizu_linear_2006} as it only imposes relative orderings.

Furthermore, we delve into another property of the path search space -- the distribution of the measure of independence from all possible causal orderings. We demonstrate that this distribution can be used to infer key graph properties such as the presence of unmeasured confounders, graph sparsity, and even the potential performance of causal discovery algorithms.

In summary, our contribution is threefold.
\begin{enumerate}
    \item \textbf{Improving LiNGAM-SPP:} We modify LiNGAM-SPP to eliminate the need for parameter tuning while yielding enhanced performance and computational efficiency.
    \item \textbf{Incorporating Known Relative Ordering:} We expand LiNGAM-SPP to incorporate prior knowledge, requiring only relative orderings for better adaptability.
    \item \textbf{Predicting Causal Graph Properties:} We develop predictive models using features derived from the path distributions of LiNGAM-SPP, specifically for predicting presence of unmeasured confounders, graph sparsity, and causal discovery algorithm performance.
\end{enumerate}

An overview of the paper is as follows. Preliminary concepts relevant to the study -- graphical causality, LiNGAM, and LiNGAM-SPP -- are contained in the second section. The succeeding three discussion sections cover the threefold contribution of the study. The modified LiNGAM-SPP and evidence of its relative superiority is laid out in the third section. The fourth section details how prior knowledge may be considered in the framework through the input and processing of known relative orderings. The section on predicting the performance of causal discovery algorithms follows, hinging on path enumeration through the use of a compact data structure called the zero-suppressed binary decision diagram (ZDD). Each of the aforementioned discussion sections presents both the methods and results, specifying employed techniques, key statistics, and relevant observations. Such an arrangement is put forward for ease of readability. The work is closed in the final section with a recapitulation if the contributions.

\section{Preliminaries}
In this section, we begin by briefly introducing graphical causality to provide context to the problem. Next, we review LiNGAM. Finally, we discuss LiNGAM-SPP and how it generalizes LiNGAM by framing it as a shortest path problem.

\subsection{Graphical Causality}
Graphical causality \cite{pearl_causality_2000, pearl_causal_2016, peters_elements_2017} uses directed acyclic graphs (DAGs) to represent causal relationships between variables. A DAG $\mathcal{G}$ can represent a joint probability distribution $P_X$ either as a factorized probability distribution or as an SCM.

The joint probability distribution $P_X$ can be represented by a factorized probability distribution according to the Markov factorization property \cite{peters_elements_2017}. Using $\mathcal{G}$, 
\begin{equation}\label{dag_factorization}
P(X_1,\ldots,X_n) = \prod_{i=1}^nP(X_i|\text{PA}^\mathcal{G}_i),
\end{equation}
where each variable $X_i$ corresponds to a node $i$ in $\mathcal{G}$ and $\text{PA}_i^\mathcal{G}$ denotes the variables corresponding to the parents of node $i$. When the edges of $\mathcal{G}$ are causal, it is referred to as a causal graph.

SCMs, on the other hand, represent $P_X$ in a slightly different way. In SCMs, each node $i$ corresponds to a variable $X_i$, a deterministic function $f_i$, an exogenous variable or noise term $U_i$ which are assumed to be jointly independent, and the probability distribution of $U_i$ which capture the stochasticity of $P_X$ \cite{pearl_causal_2016, scholkopf_toward_2021}. This is given by 
\begin{equation} \label{scm_form}
    X_i := f_i(\text{PA}_i, U_i).
\end{equation}

SCMs provide a more expressive representation than causal graphs by explicitly modeling the causal relationships using deterministic functions. In Pearl's Ladder of Causation \cite{pearl_book_2018}, SCMs are placed at a higher level than causal graphs. While both can represent interventions, only SCMs can capture counterfactual reasoning. Regardless, both of these representations require knowing the structure of the underlying DAG $\mathcal{G}$, which can be obtained through causal discovery methods. One of such methods is LiNGAM, which is discussed in the following section.

\subsection{LiNGAM}
LiNGAM \cite{shimizu_linear_2006, shimizu_directlingam_2011, shimizu_lingam_2014} assumes an SCM with linear functions, non-Gaussian noise terms, and no unmeasured confounders. Specifically, this is represented as linear equations in matrix form 
\begin{equation} \label{lingam_mat_form}
	\textbf{x}= \textbf{Bx} + \textbf{e},
\end{equation}
where $\textbf{B}$ is a strictly lower triangular adjacency matrix and the elements of $\textbf{e}$ are continuous non-Gaussian distributions with zero mean and nonzero variance. The strictly lower triangular adjacency matrix implies that each variable is expressed as the linear combination of the variables that precede it plus the noise term. In effect, LiNGAM aims to discover the causal ordering of the variables. Once the causal ordering is obtained, it is possible to determine which connections are zero using sparse regression methods \cite{shimizu_lingam_2014, zou_adaptive_lasso_2006}.

The original LiNGAM paper \cite{shimizu_linear_2006} uses independent components analysis (ICA) to estimate Eq. \eqref{lingam_mat_form}, which is why it is often called ICA-LiNGAM in literature \cite{glymour_review_2019,spirtes_causal_2016, shimizu_directlingam_2011, shimizu_lingam_2014}. A shortcoming of ICA-LiNGAM, however, is that it does not guarantee convergence to a correct solution in a finite number of steps. To address this, DirectLiNGAM \cite{shimizu_directlingam_2011} was proposed.

DirectLiNGAM provides a causal ordering in the same number of steps as the number of variables. In each step, DirectLiNGAM selects the variable that is the most independent -- with the smallest mutual information -- from the least squares residuals of the remaining unselected variables. Subsequently, DirectLiNGAM removes the effect of the selected variable by reassigning the remaining variables with the residuals from the least squares. This process iterates until the causal order is obtained \cite{shimizu_directlingam_2011}.

DirectLiNGAM allows for various measures of independence \cite{shimizu_directlingam_2011}. While the original implementation uses a kernel-based mutual information estimator \cite{shimizu_directlingam_2011}, it also discusses alternative measures such as the Hilbert-Schmidt independence criterion (HSIC) \cite{gretton_2007}, a $k$NN-based mutual information estimator \cite{kraskov_estimating_2004}, and single nonlinear correlation \cite{hyvarinen_1997}. Moreover, a measure known as the pairwise likelihood ratio (PLR) is introduced, with its efficacy as an independence measure for DirectLiNGAM demonstrated \cite{hyvarinen13a}. Despite its computational simplicity, PLR performs at least as well and sometimes even better than ICA-LiNGAM and kernel-based DirectLiNGAM, especially in cases with a limited number of data points relative to the data dimension or when the data is noisy. The computational efficiency of PLR also renders it significantly faster than the kernel-based DirectLiNGAM \cite{hyvarinen13a}. As of writing, the official \texttt{lingam} package \cite{lingam_package_2023} defaults to using PLR in its implementation of DirectLiNGAM.

\subsection{LiNGAM-SPP}
LiNGAM-SPP was introduced as a generalization of DirectLiNGAM \cite{suzuki_lingam_mmi}. Unlike DirectLiNGAM, which determines variable ordering by selecting the most independent feature at each step, LiNGAM-SPP seeks the ordering that minimizes the total mutual information across all steps. The authors of LiNGAM-SPP made the assumption, supported through simulations, asserting that LiNGAM-SPP can accurately determine the causal order even in the presence of unmeasured confounders \cite{suzuki_lingam_mmi}. This characteristic makes LiNGAM-SPP a generalization of DirectLiNGAM, as the latter assumes the absence of unmeasured confounders.

\begin{figure}[h]
  \centering
  \includegraphics[width=0.4\linewidth]{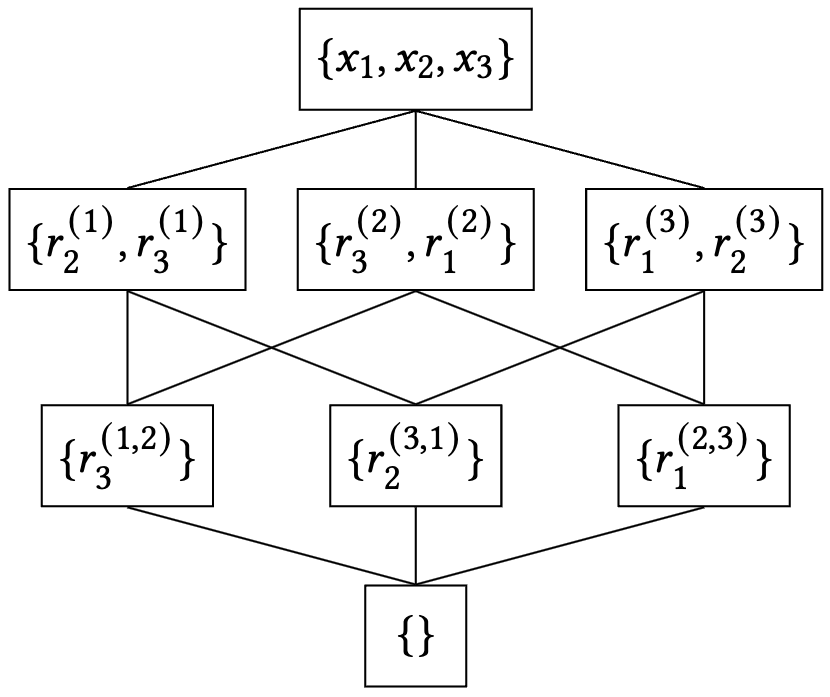}
  \caption{LiNGAM-SPP: A shortest-path formulation of the causal ordering problem. Adapted from the LiNGAM-SPP paper \cite{suzuki_lingam_mmi} with slight changes in notation.}
  \Description{The figure depicts how LiNGAM-SPP frames the causal ordering problem as a shortest path problem for the three-variable case. Nodes correspond to candidate variables and edges signify some independence measure. The objective is to find the shortest path from the initial node $\{x_1, x_2, x_3\}$, where all variables are still candidates, to the terminal node $\{\}$, which indicates that the variables have been ordered.}
\label{fig:lingam_mmi_graph}
\end{figure}

LiNGAM-SPP formulates the causal ordering problem as a graph search problem \cite{suzuki_lingam_mmi}. Figure \ref{fig:lingam_mmi_graph} shows an example of how this is done for the three-variable case. In this graph, nodes represent candidate variables, edges represent some measure of independence, and the goal is to find a path from the starting node $\{x_1, x_2, x_3\}$ to the goal node $\{\}$. The selected variable and its independence can be inferred from the subscripts of candidate variables. For instance, the edge between $\{x_1, x_2, x_3\}\rightarrow\{r_2^{(1)}, r_3^{(1)}\}$ represents the mutual information if $x_1$ were selected as the first variable and its edge weight is the independence of $x_1$ vis-\`{a}-vis the residuals. Mathematically, this mutual information is denoted by $I(x_1, \{r_2^{(1)}, r_3^{(1)}\})$, where the residual is defined as
\begin{equation} \label{res_def}
    r_i^{(j)} = x_i - \frac{cov(x_i,x_j)}{var(x_j)} x_j.
\end{equation}
Here, $r_i^{(j)}$ represents the residual when $x_i$ is regressed on $x_j$. Similar to DirectLiNGAM, these residuals are stored for use in the next stage. For example, the edge $\{r_2^{(1)}, r_3^{(1)}\} \rightarrow \{r_3^{(1,2)}\}$ has an edge weight of $I(r_2^{(1)}, r_3^{(1,2)})$, the mutual information if $x_2$ were the second variable chosen after $x_1$, where $r_3^{(1,2)}$ is the residual when $r_3^{(1)}$ is regressed on $r_2^{(1)}$.

The path formulation is made possible by the additive property of mutual information \cite{kraskov_estimating_2004}. In notation,
\begin{equation}
    I(X,Y,Z) = I(\{X,Y\}, Z) + I(X,Y).
\end{equation}
This property facilitates the step-wise decomposition and calculation of mutual information to obtain the total mutual information. In the context of LiNGAM-SPP, we use this property to minimize $I(\epsilon_1, \epsilon_2, \ldots, \epsilon_p)$, where $\epsilon_i$ represents the noise terms of the variables according to Eq. \ref{lingam_mat_form}. We minimize this to satisfy the SCM property that the noise terms should be jointly independent \cite{pearl_causal_2016, scholkopf_toward_2021}. Referring to the example from Figure \ref{fig:lingam_mmi_graph}, if the causal order is $x_1, x_2, x_3$, then the total mutual information is
\begin{equation}
    I(\epsilon_1, \epsilon_2, \epsilon_3) = I(x_1, \{r_2^{(1)}, r_3^{(1)}\}) + I(r_2^{(1)}, r_3^{(1,2)}) + 0.
\end{equation}
This is the sum of the edge weights along the path $\{x_1, x_2, x_3\}\rightarrow\{r_2^{(1)}, r_3^{(1)}\}\rightarrow\{r_3^{(1,2)}\}\rightarrow \{\}$, assuming all edges connecting to the $\{\}$ node have a value of 0, \textit{i.e.}, $I(r_i^{(j)}, \{\}) = 0$. 

Finally, in order to look for the shortest path, LiNGAM-SPP uses Dijkstra's algorithm with lazy evaluation. The mutual information or the edge weights are only computed when needed. Please refer to either LiNGAM-SPP paper \cite{suzuki_lingam_mmi, suzuki2024generalization} for a more detailed description of the algorithm.

\section{Improving LiNGAM-SPP}
In the initial LiNGAM-SPP paper, it was demonstrated that LiNGAM-SPP outperforms DirectLiNGAM, specifically when comparing the $k$NN-based LiNGAM-SPP and the HSIC-based DirectLiNGAM \cite{suzuki_lingam_mmi}. A succeeding paper \cite{suzuki2024generalization} then showed that the LiNGAM-SPP based on copula entropy (copent) outperforms different implementations of DirectLiNGAM -- PLR \cite{hyvarinen13a}, kernel \cite{shimizu_directlingam_2011}, HSIC \cite{hyvarinen13a}, and ICA \cite{shimizu_linear_2006}. While this initial success is promising, relying on the $k$NN-based mutual information raises concerns because it requires tuning the parameter $k$ for optimal performance. Unfortunately, this parameter tuning is not feasible in practical applications where the true data generating process is unknown.

To better test the performance of the method, we aim to test LiNGAM-SPP on a more general class of simulations and on real-world data. Given the above objectives, we introduce the following improvements on which the subsections are based.
\begin{enumerate}
    \item We adopt the data generating process used in the DirectLiNGAM paper \cite{shimizu_directlingam_2011} with a few additions for simulating confounders. This process is more general than the one used in the previous LiNGAM-SPP papers, providing a broader scope for testing and evaluation.
    \item Instead of the $k$NN-based mutual information estimator \cite{kraskov_estimating_2004} for LiNGAM-SPP, we propose adopting the PLR as the independence measure \cite{hyvarinen13a}. This modification not only enhances performance and computation time but also eliminates the need for fine-tuning.
    \item To validate the effectiveness of the method in practical settings, we conduct tests on LiNGAM-SPP using real-world data and compared its results with DirectLiNGAM. This step provides insights into the applicability of the method beyond simulated scenarios.
\end{enumerate}

\subsection{Data Generating Process}
Our data generating process was adapted from \cite{shimizu_directlingam_2011}, which we extended to incorporate cases with unmeasured confounders. The framework outlined in \cite{tashiro_parcelingam_2013, hoyer_nonlinear_2008} is followed, where LiNGAM with unmeasured confounders can be represented as
\begin{equation}
    \mathbf{x} = \mathbf{Bx} + \boldsymbol{\Lambda} \mathbf{f} + \mathbf{e}.
\end{equation}

In the equation above, $\mathbf{x}$ represents the observed data, $\mathbf{B}$ is a $p \times p$ matrix where $p$ is the number of features, $\mathbf{f}$ is the latent confounder vector, and $\mathbf{e}$ is the non-Gaussian noise. The matrix $\boldsymbol{\Lambda}$, with elements $\lambda_{ij}$, denotes the connection strength between $f_j$ and $x_i$. Additionally, each unmeasured confounder should have at least 2 children and the matrix $\boldsymbol{\Lambda}$ should be of full column rank \cite{tashiro_parcelingam_2013, hoyer_nonlinear_2008}. To implement this, we perform the following steps.
\begin{enumerate}
    \item We initialize the matrix $\mathbf{B}$ by creating a $p \times p$ lower triangular matrix with elements randomly drawn from a uniform distribution in the interval $[-1.5, -0.5] \cup [0.5, 1.5]$, following \cite{silva_learning_2006a,shimizu_directlingam_2011}.
    \item To introduce sparsity into the graph, we applied an element-wise product with a Bernoulli distribution matrix, following the approach of \cite{shimizu_directlingam_2011,kalisch_estimating_2007a}. The success probability parameter of this Bernoulli distribution determines the sparsity of the graph.
    \item For modeling non-Gaussian noise and confounders, we used the set of 18 non-Gaussian distributions used in \cite{bach_kernel_2003,shimizu_directlingam_2011}. Each noise term and confounder were randomly sampled from this set. Moreover, we scaled these distributions to ensure that their variances fell within the interval $[1, 3]$, as in \cite{silva_learning_2006a,shimizu_directlingam_2011}. We also included a scaling parameter for the confounders which we refer to as confounding strength.
    \item To construct $\boldsymbol{\Lambda}$, we generated its columns sequentially. For each column to be added, we first selected two variables to be influenced. Subsequently, we created a binary column vector by sampling it from a Bernoulli distribution where the probability of success, an adjustable parameter, may be interpreted as the confoundedness of the graph. The two selected variables and the binary vector are combined to create a vector determining which features are to be confounded. In effect, each column of $\boldsymbol{\Lambda}$ will have at least 2 nonzero elements, even if the confoundedness is set to 0. Before appending the column to $\boldsymbol{\Lambda}$, we make sure that it is not equal to any of the existing columns, hence guaranteeing that $\boldsymbol{\Lambda}$ remains full rank.
\end{enumerate}

Our data generating process is designed to capture a broad range of linear non-Gaussian processes. It allows us to control various parameters such as the number of features $p$, sample size $N$, sparsity, confoundedness, confounding strength, and the number of confounders. In the following sections where we present results of simulations, the simulations used sparsity drawn from a uniform distribution $\text{U}(0,1)$, confounding strength $10^s$ with $s$ drawn from $\text{U}(1,2)$, the number of confounders randomly drawn from $\{1,2,3\}$, and confoundedness drawn from $\text{U}(0,1)$.

\subsection{PLR as an Independence Measure}
The initial LiNGAM-SPP paper \cite{suzuki_lingam_mmi} demonstrated that using the $k$NN-based mutual information estimator \cite{kraskov_estimating_2004} with LiNGAM-SPP led to superior performance compared to the HSIC-based DirectLiNGAM. However, a limitation was that the parameter $k$ had to be tuned to achieve optimal performance. A subsequent study by the same authors \cite{suzuki2024generalization} used copent \cite{ma2011_copent} as the mutual information estimator for LiNGAM-SPP which, while achieving commendable performance, still relied on the $k$NN-based mutual information estimator \cite{kraskov_estimating_2004}. In this study, we propose using the PLR as an alternative to the $k$NN-based mutual information estimator. Notably, PLR eliminates the need for parameter tuning, demonstrates superior performance, and is computationally more efficient.

PLR uses the likelihood ratio, the difference between the log-likelihoods, to determine the causal direction between two non-Gaussian variables. For instance, in determining the causal direction between $x$ and $y$, both standardized to have a mean of 0 and a variance of 1, PLR computes the difference between the log-likelihoods of both causal directions
\begin{equation}
    R = \frac{1}{T}\log L(x\rightarrow y) - \frac{1}{T}\log L(y \rightarrow x).  
\end{equation}
Here, the log-likelihoods are normalized by the sample size $T$. Given $R$, we can infer the correct direction. If $R$ is positive, then $x \rightarrow y$; otherwise, $y \rightarrow x$. If $x\rightarrow y$, the SCM is $y=\rho x + d$, where $d \perp\!\!\!\perp x$. On the other hand, if $y \rightarrow x$, then the SCM is $x = \rho y + e$, where $e \perp\!\!\!\perp y$. Note that the coefficient $\rho$ is the same for both directions as it represents the correlation coefficient. Normally, computing $R$ requires knowing the distributions of $x$, $y$, $d$, and $e$. A practical approximation using differential entropy \cite{hyvarinen13a} was nonetheless derived to compute this without prior knowledge of the distributions. Mathematically, 
\begin{equation}
    R \longrightarrow -H(x) - H(\hat{d}/\sigma_d) + H(y) + H(\hat{e}/\sigma_e).
\end{equation}
Here, $\hat{d} = y - \rho x$ and $\hat{e} = x - \rho y$ represent the estimated residuals and $\sigma_d$ and $\sigma_e$ denote their standard deviations. The closed-form of the differential entropy $H$ may be approximated by
\begin{equation}
    \hat{H}(u) = H(v) - k_1[E\{\log\cosh u\} - \gamma]^2 - k_2[E\{u \exp (-u^2/2)\}]^2,
\end{equation}
where $H(v) = \frac{1}{2}(1+\log 2\pi)$ and the constants have numerical values $k_1\approx 79.047$, $k_2 \approx 7.4129$, and $\gamma \approx 0.37457$ \cite{hyvarinen1998_diffentropy}. This closed-form approximation contributes significantly to the efficiency of PLR.

In our implementation, we incorporate code from the \texttt{lingam} package \cite{lingam_package_2023}, which currently defaults to using PLR in its DirectLiNGAM implementation. We use the relevant portions of this code to integrate the PLR measure into the LiNGAM-SPP framework with slight modifications. Specifically, the original implementation calculated the PLR independence measure $m_i$ for each feature $i$ as
\begin{equation}
    m_i = -\sum_j \min(0, [M]_{ij})^2,
\end{equation}
where $[M]_{ij}$ is the PLR between feature $i$ and $j$ \cite{hyvarinen13a}. We introduce two modifications to adapt it to the shortest-path framework. First, we remove the negative sign and reformulated it as a minimization problem rather than a maximization one. Second, we normalize $m_i$ by dividing it by the number of terms in the summation. This adjustment is crucial because failing to do so could result in earlier steps having larger values of $m_i$ because they involve more terms. Such an imbalance in path measurements might pose challenges in identifying the shortest path.

As previously discussed, we propose using the PLR as the mutual information estimator for LiNGAM-SPP, which we refer to as PLR LiNGAM-SPP. To evaluate the performance of PLR LiNGAM-SPP, we conducted experiments on a data generating process adapted from the DirectLiNGAM paper \cite{shimizu_directlingam_2011}, with modifications to incorporate simulated unmeasured confounders. This data generating process differs from, but is more comprehensive than, the one utilized in the original LiNGAM-SPP studies \cite{suzuki_lingam_mmi, suzuki2024generalization}, so the results here may be different from these studies. Furthermore, we assessed the efficacy of the method in practical settings by testing it on real-world datasets, comparing the results with those obtained from DirectLiNGAM. This validation demonstrates the utility of the method beyond simulated scenarios.

To showcase the effectiveness of this enhancement, we compare our proposed method, PLR LiNGAM-SPP, with other existing methods -- PLR DirectLiNGAM \cite{hyvarinen13a}, $k$NN LiNGAM-SPP \cite{suzuki_lingam_mmi}, and copent LiNGAM-SPP \cite{suzuki2024generalization}. Following the methodology in the original $k$NN LiNGAM-SPP paper \cite{suzuki_lingam_mmi}, we used\cite{kraskov_estimating_2004}
\begin{equation}
    I(X,Y) = \psi(k) - \big<\psi(n_x+1) + \psi(n_y+1)\big> + \psi(N),
    \label{eq:knn_mi}
\end{equation}
where $\psi(.)$ represents the digamma function, as the definition for the $k$NN-based mutual information estimator. To implement this, we adapted code from the feature selection module of the \texttt{sklearn} \cite{scikit-learn} library. We extend it to accomodate a one-to-many variable setting, \textit{i.e.}, $I((X_1,X_2,\ldots , X_{m-1}), X_m)$, by redefining $n_x$ in Eq. \ref{eq:knn_mi} to represent the number of points in the $(m-1)$-dimensional space as outlined by \cite{kraskov_estimating_2004}. Meanwhile, we use the Python package \texttt{copent} \cite{ma2011_copent} for computing the copent in copent LiNGAM-SPP. Finally, performance was evaluated using the same error metric as previous LiNGAM-SPP papers \cite{suzuki_lingam_mmi, suzuki2024generalization} where the ordering error, denoted as $E_o$, is defined by
\begin{equation}
    E_o = \frac{2r}{p(p-1)},
\end{equation}
where $p$ represents the number of features and $r$ is the count of pairs in the wrong order. This essentially counts the fraction pairs in the incorrect order. 

As previously mentioned, both $k$NN and copent LiNGAM-SPP require the tuning of the $k$ parameter, which is not possible in practical settings. However, for the purpose of comparison, we set $k$ to 5\%, 10\%, and the square root of the sample size ($N$) instead of parameter tuning. We then select the best-performing configuration. Table \ref{tab:criterionB_comparison} presents results for various scenarios. The reported values represent the average ordering errors, $E_o$, across 250 trials for each scenario. The results and conclusions drawn from this comparison may differ from the previous LiNGAM-SPP studies since a different, albeit more general, data generating process was used for the simulations. 

\begin{table}
  \caption{Average Ordering Errors $E_o$ Across Various Methods and Scenarios}
  \label{tab:criterionB_comparison}
  \begin{tabular}{rr|rrrr|rrrr}
\toprule
   \multicolumn{2}{c}{}& \multicolumn{4}{c}{Without Confounders} & \multicolumn{4}{c}{With Confounders} \\
\hline
   & Sample Size (N) &  100  &  250  &  500  &  1000 &  100  &  250  &  500  &  1000 \\
p & Method &       &       &       &       &       &       &       &       \\
\midrule
5  & PLR LiNGAM-SPP &  0.29 &  0.28 &  0.23 &  0.21 &  0.46 &  0.45 &  0.42 &  0.39 \\
   & PLR DirectLiNGAM &  0.29 &  0.28 &  0.23 &  0.22 &  0.46 &  0.45 &  0.42 &  0.40 \\
  & Copent LiNGAM-SPP &  0.40 &  0.41 &  0.38 &  0.39 &  0.43 &  0.40 &  0.43 &  0.42 \\
   & $k$NN LiNGAM-SPP &  \textbf{0.29} &  \textbf{0.23} &  \textbf{0.22} &  \textbf{0.18} &  \textbf{0.40} &  \textbf{0.37}&  \textbf{0.35} &  \textbf{0.33} \\
\hline
10 & PLR LiNGAM-SPP &  0.33 &  \textbf{0.24} &  \textbf{0.21} &  \textbf{0.18} &  0.44 &  \textbf{0.40} &  \textbf{0.37} &  \textbf{0.34} \\
   & PLR DirectLiNGAM &  \textbf{0.32} &  0.25 &  0.22 &  0.19 &  \textbf{0.44} &  0.41 &  0.37 &  0.36 \\
   & Copent LiNGAM-SPP &  0.49 &  0.48 &  0.50 &  0.49 &  0.46 &  0.46 &  0.45 &  0.46 \\
   & $k$NN LiNGAM-SPP &  0.48 &  0.40 &  0.35 &  0.30 &  0.49 &  0.43 &  0.39 &  0.35 \\
\hline
15 & PLR LiNGAM-SPP &  0.33 &  \textbf{0.24} &  \textbf{0.19} &  \textbf{0.16} &  \textbf{0.41} &  \textbf{0.36} &  0.35 &  \textbf{0.29} \\
   & PLR DirectLiNGAM &  \textbf{0.32} &  0.25 &  0.21 &  0.18 &  0.42 &  0.37 &  \textbf{0.33} &  0.32 \\
  & Copent LiNGAM-SPP &  0.51 &  0.50 &  0.51 &  0.52 &  0.48 &  0.48 &  0.49 &  0.49 \\
   & $k$NN LiNGAM-SPP &  0.56 &  0.51 &  0.45 &  0.38 &  0.54 &  0.49 &  0.46 &  0.40 \\
\bottomrule
\end{tabular}

\end{table}

While $k$NN LiNGAM-SPP excels for $p=5$, PLR LiNGAM-SPP outperforms in most cases for larger $p$ with DirectLiNGAM closely trailing. A paired t-test reveals that overall, PLR LiNGAM-SPP shows a statistically significant improvement over DirectLiNGAM with a p-value of $5.22 \times 10^{-5}$.

Interestingly, the performance disparity between PLR LiNGAM-SPP and DirectLiNGAM widens as the number of features $p$ increases. We hypothesize that LiNGAM-SPP demonstrates greater robustness particularly with longer sequences, owing to its ability for backtracking and correction. In contrast to DirectLiNGAM which is prone to error propagation -- wherein a mistake in one step compromises subsequent steps, LiNGAM-SPP utilizing Dijkstra's algorithm exhibits a mechanism for error correction. If an incorrect feature is selected in one step, subsequent steps are more likely to yield large mutual information values. In such instances, Dijkstra's algorithm can backtrack to explore alternative paths with smaller overall values.

\begin{table}
  \caption{Average Runtime Across Various Methods and Scenarios}
  \begin{tabular}{rr|rrrr|rrrr}
\toprule
\multicolumn{2}{c}{} & \multicolumn{4}{c}{Without Confounders} & \multicolumn{4}{c}{With Confounders} \\
\hline
  & Sample Size ($N$) &    100  &   250  &   500  &    1000   &   100  &   250  &   500  &    1000 \\
p & Method &       &         &        &        &        &        &        &         \\
\midrule
5   & PLR LiNGAM-SPP &  0.44s &  0.44s &  0.45s &     0.46s &  0.43s &  0.45s &  0.44s &   0.46s \\
   & PLR DirectLiNGAM &  0.04s &  0.04s &  0.04s &     0.05s &  0.04s &  0.04s &  0.04s &   0.05s \\
  & Copent LiNGAM-SPP &  0.06s &  0.11s &   0.3s &      1.1s &  0.07s &  0.13s &  0.35s &    1.3s \\
   & $k$NN LiNGAM-SPP &  0.27s &  0.39s &  0.72s &      1.8s &  0.26s &  0.37s &  0.65s &    1.7s \\
\hline
10   & PLR LiNGAM-SPP &   6.8s &   6.9s &   6.9s &      6.1s &   6.4s &   7.2s &     7s &    7.6s \\
   & PLR DirectLiNGAM &  0.26s &  0.27s &   0.3s &     0.36s &  0.26s &  0.27s &  0.31s &   0.36s \\
 & Copent LiNGAM-SPP &   0.4s &   0.6s &   1.3s &      4.6s &  0.48s &   0.7s &   1.6s &    5.5s \\
   & $k$NN LiNGAM-SPP &    32s &    45s &   1.4m &      3.5m &    32s &    44s &   1.3m &    3.4m \\
\hline
15  & PLR LiNGAM-SPP &   2.2m &   1.6m &   1.2m &      1.5m &   1.4m &   3.6m &   2.8m &    3.2m \\
   & PLR DirectLiNGAM &  0.83s &   0.9s &  0.98s &      1.2s &  0.84s &  0.92s &     1s &    1.2s \\
 & Copent LiNGAM-SPP &   1.3s &   1.8s &   3.4s &       11s &   1.6s &   2.1s &   4.3s &     14s \\
   & $k$NN LiNGAM-SPP &    37m &    48m &    84m &  206m &    36m &    49m &    85m &  205m \\
\bottomrule
\end{tabular}

  \label{tab:time_comparison}
\end{table}

Furthermore, Table \ref{tab:time_comparison} presents the average run times for the same cases. As expected, DirectLiNGAM, being a greedy algorithm, exhibits the fastest runtimes. Meanwhile, the runtimes of copent and $k$NN LiNGAM-SPP, both of which rely on the $k$NN mutual information estimator, tend to increase with sample size $N$. Despite this trend, copent LiNGAM-SPP has the smallest runtimes out of all LiNGAM-SPP methods. On the other hand, $k$NN LiNGAM-SPP runs notably slower. For instance, at $p=15$, the runtime can extend to hours. While copent LiNGAM-SPP is faster, its performance remains subpar. Overall, PLR LiNGAM-SPP presents the most favorable compromise between runtime and performance, with the added advantage of not requiring parameter tuning.

\subsection{Testing on Real-World Data}
To evaluate PLR LiNGAM-SPP on real-world data, we sourced datasets from the Carnegie Mellon University Causal Learning and Reasoning Group list of benchmarks accessible on GitHub (\url{https://github.com/cmu-phil/example-causal-datasets}). Our criteria for selection included datasets with continuous features and known ground truth. Additionally, we restricted our choices to datasets with $p \leq 20$ due to the computational demands of the LiNGAM-SPP method for larger values of $p$.

To derive the causal graph from LiNGAM-SPP, we adapted code from the \texttt{lingam} package \cite{lingam_package_2023}. Specifically, after getting the causal order, we used the same sparse regression as DirectLiNGAM to identify which edges are zero \cite{zou_adaptive_lasso_2006, shimizu_directlingam_2011}. Table \ref{tab:real_data_performance} provides the ground truth for each dataset, indicating required and forbidden edges, along with the edges captured by PLR DirectLiNGAM and PLR LiNGAM-SPP. The ground truth originally included tiers, which we converted into forbidden edges given that later tiers cannot cause earlier ones. For more details about the datasets and the provided ground truth, please refer to the GitHub repository.

\begin{table}
  \centering
  \caption{Performance of PLR DirectLiNGAM and PLR LiNGAM-SPP on Real-World Datasets}
  \begin{tabular}{rc|cc|cc|cc}
\toprule
        \multicolumn{2}{c}{}  &  \multicolumn{2}{c}{Known Edges} &  \multicolumn{2}{c}{DirectLiNGAM} & \multicolumn{2}{c}{LiNGAM-SPP} \\
\hline
                Dataset &  p & Req. &  Forb. &  Req. &  Forb. &  Req. &  Forb. \\
\hline
               sachs \cite{data_sachs}&            11 &              20 &                0 &                     10 &                       0 &                   10 &                     0 \\
 yacht-hydrodynamics \cite{data_yacht_hydrodynamics} &             7 &               1 &               15 &                      0 &                       2 &                    0 &                     2 \\
             abalone \cite{data_abalone} &             8 &               0 &               17 &                      0 &                       4 &                    0 &                     4 \\
  airfoil-self-noise \cite{data_airfoil_self_noise} &             6 &               0 &               11 &                      0 &                       6 &                    0 &                     6 \\
    wine-quality-red \cite{data_wine_quality}&            12 &               2 &               11 &                      1 &                       7 &                    2 &                     2 \\
  wine-quality-white \cite{data_wine_quality} &            12 &               2 &               11 &                      2 &                       4 &                    2 &                     1 \\
\bottomrule
\end{tabular}

  \label{tab:real_data_performance}
\end{table}

The results in Table \ref{tab:real_data_performance} reveal that for the wine-quality-red and wine-quality-white datasets, PLR LiNGAM-SPP outperforms PLR DirectLiNGAM, capturing fewer forbidden edges and more required edges. For the other datasets, both methods yield identical results. This observation aligns with the findings in Table \ref{tab:criterionB_comparison}, indicating that PLR LiNGAM-SPP is more robust for longer sequences. This also suggests that PLR LiNGAM-SPP performs at least as well as PLR DirectLiNGAM in most cases. 

\section{Incorporating Known Relative Ordering}
One notable feature of DirectLiNGAM is its capability to incorporate prior knowledge to enhance overall results by allowing users to specify which directed edges are required, forbidden, or unknown \cite{shimizu_directlingam_2011}. In this study, we introduce an enhancement to LiNGAM-SPP to allow for integrating such prior knowledge. Unlike DirectLiNGAM, where the exact edges need to be specified, our implementation only requires specifying the relative order. In practical terms, this means we can assert that one feature causes another while still allowing for the possibility of having intermediary features, or mediators, between them.

To implement this, we modify the shortest path algorithm to selectively skip nodes that violate the specified order. The path formulation of LiNGAM-SPP naturally lends itself to this task. Each node in the graph representation of causal orderings contains information about relative orderings, indicating which features have been selected and which are yet to be explored. Specifically, if a feature is absent from the subscripts, it is positioned earlier in the causal order. For instance, referencing the graph in Figure \ref{fig:lingam_mmi_graph}, if we assert that $x_1$ precedes $x_2$, then nodes with subscript ${1}$ but not ${2}$ -- such as $\{r_3^{(2)}, r_1^{(2)}\}$ and $\{r_1^{(2,3)}\}$ -- should be skipped. This is because they imply that $x_2$ comes before $x_1$, violating the known ordering.

\begin{figure}[h]
  \centering
  \includegraphics[width=0.8\linewidth]{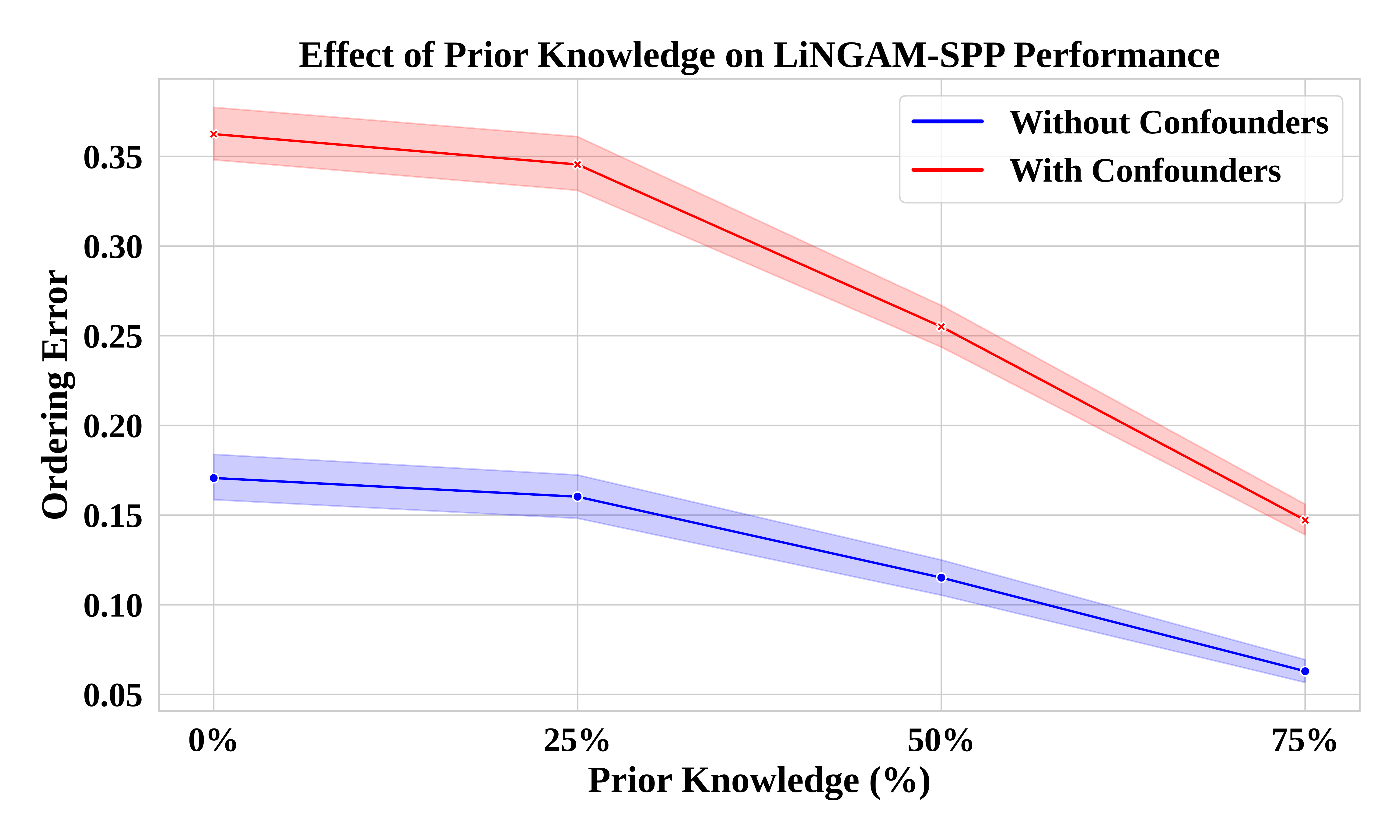}
    \caption{Performance Impact of Prior Knowledge: Combined results for $p = 4,8,12$.}
  \Description{This figure shows a decreasing trend in the ordering error as the percentage of incorporated prior knowledge increases. It presents the combined results for different number of features (p = 4,8, and 12).}
\label{fig:prior_knowledge_performance}
\end{figure}

We demonstrate the impact of incorporating prior knowledge on performance in Figure \ref{fig:prior_knowledge_performance} and Table \ref{tab:prior_knowledge_edges}. To quantify the extent of prior knowledge, we express it as a percentage of variables with specified orderings. For example, if $p = 8$, 50\% knowledge means that the relative ordering of 4 variables is given. In this demonstration, the input is the relative ordering of 2 or more variables, which we then convert to variable pairs. As illustration, $(1,2,3)$ implies $(1,2)$, $(2,3)$, $(1,3)$. It is, however, also possible to manually specify pairs. As expected, incorporating more prior knowledge improves performance and generally reduces the traversed number of edges.

\begin{table}
  \caption{Computed Edges with Prior Knowledge}
  \begin{tabular}{r|rrrr|rrrr}
\toprule
 \multicolumn{1}{c}{}& \multicolumn{4}{l}{Without Confounders} & \multicolumn{4}{l}{With Confounders} \\
 \hline
 Prior Knowledge & 0\% & 25\% & 50\% & 75\% & 0\% & 25\% & 50\% & 75\% \\
p  &                            &                             &                             &                             &                            &                             &                             &                             \\
\midrule
4  &                          9 &                           9 &                           8 &                           6 &                          9 &                           9 &                           8 &                           6 \\
8  &                         35 &                          34 &                          27 &                          18 &                         39 &                          40 &                          37 &                          22 \\
12 &                        107 &                          99 &                          81 &                          42 &                        184 &                         198 &                         174 &                          74 \\
\bottomrule
\end{tabular}

  \label{tab:prior_knowledge_edges}
\end{table}

Relative orderings provide a more flexible way of incorporating prior knowledge, which is valuable because we do not always know the exact edges. It is more common for us to know the relative ordering rather than the exact edges. For instance, in most applications, it is known that nothing can affect the age, sex, and other background variables of a person so these must be positioned earlier in the ordering relative to other variables. Another example is incorporating temporal information, where features with time-related attributes may exist and we only know the temporal order but not necessarily the ordering among the features within the same time slice.

\section{Predicting Causal Graph Properties}
In the preceding sections, the advantage of LiNGAM-SPP, which reframes the causal ordering problem as a shortest path problem, combined with original enhancements were discussed. However, we find that the utility of this formulation of LiNGAM extends beyond finding the shortest path. We discover that generating the distribution of the total path lengths from all possible orderings through path enumeration reveals crucial insights into the causal graph properties. Particularly, we observe that the path distribution may be used to infer graph properties such as the presence of unmeasured confounders, the sparsity of the graph, and to some extent, the reliability of PLR DirectLiNGAM and PLR LiNGAM-SPP. The presence of unmeasured confounders, to emphasize, is an essential assumption to verify especially for LiNGAM.

\begin{figure}[h]
    \centering
    \includegraphics[width=\linewidth]{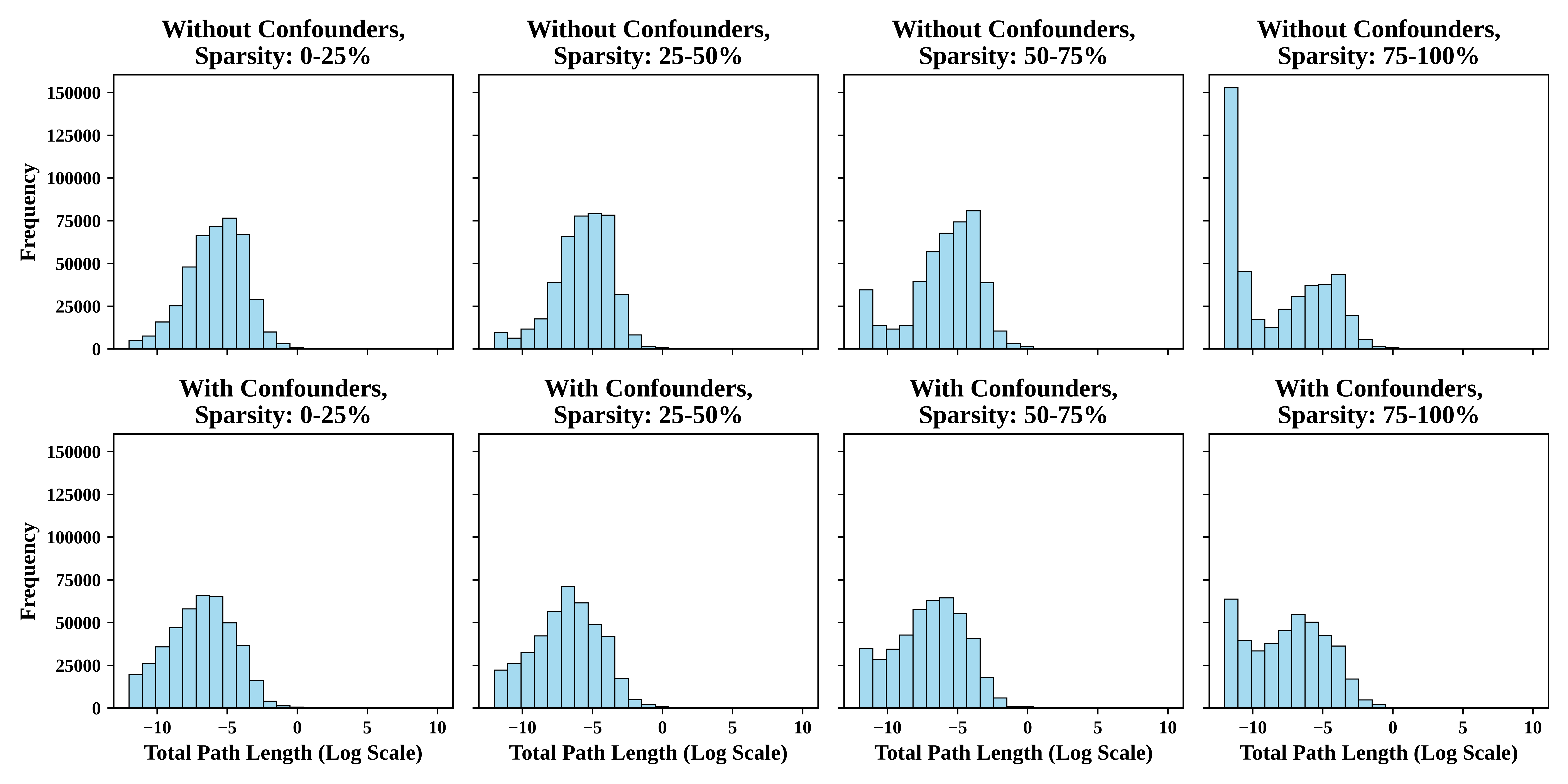}
    \caption{Log-transformed Path Distributions for Various Scenarios}
    \Description{The figure presents 8 subfigures depicting log-transformed path distributions across various scenarios. These scenarios include cases with and without confounders, each representing 4 levels of sparsity: 0-25\%, 25-50\%, 50-75\%, and 75-100\%. Notably, discernible differences in distribution shapes emerge among the scenarios. In more challenging instances, such as those involving confounders or higher graph sparsity, there is an observable increase in the frequency of shorter paths. This suggests that determining the shortest path becomes progressively more difficult and ambiguous under these conditions.}
    \label{fig:path_enumeration_hist_eda}
\end{figure}

Our inspiration stems from the observations in Figure \ref{fig:path_enumeration_hist_eda}. These histograms illustrate the combined log-transformed PLR path distributions for $p=4,5,6$ using the same data generating process and simulation configurations as the previous sections. Notably, clear patterns emerge in the path distributions for the different cases. More challenging scenarios such as the presence of confounders or increased graph sparsity show increased frequency of shorter paths. This may indicate that correct ordering becomes more challenging and ambiguous in these cases. 

To take advantage of the insight that the path distribution informs us about graph properties, we extract standardized moments to capture the shape of the distribution. Specifically, we compute the standardized moments of the log-transformed path lengths. These standardized moments will then serve as features in models trained to predict the various graph properties. The standardized moment is defined as
\begin{equation}
    \tilde{\mu}_k = \frac{\text{E}[(X - \mu)^k]}{(\text{Var}[X-\mu])^{k/2}}.
\end{equation}
Here, \(\tilde{\mu}_k\) denotes the \(k\)-th standardized moment and \(X\) represents the log-transformed path lengths. To test the generalizability of the models, \(p=4,5,6\) were used for training while \(p=7\), unseen in the training set, is used for testing. 

In the subsequent sections, we form a suite of classical machine learning algorithms. The $k$NN \cite{scikit-learn}, Random Forest \cite{scikit-learn}, XGBoost \cite{chen_xgboost_2016}, CatBoost \cite{dorogush2018catboost}, and AdaBoost \cite{scikit-learn}) are used to train the following models.
\begin{enumerate}
    \item \textbf{Confounder Detector --} a classifier to detect the presence of unmeasured confounders.
    \item \textbf{Sparseness Estimator --} a classifier and a regression model to estimate graph sparseness.
    \item \textbf{Performance Predictors --} classifier and regression models to predict the reliability of causal discovery algorithms.
\end{enumerate}
We use the receiver operating characteristic (ROC) curve as basis for the metric of discrimination, the area under the ROC curve (AUC).

\subsection{Path Enumeration via ZDDs}
Before delving into the details of the models, this subsection highlights our decision to use ZDDs for path enumeration. ZDDs are compact data structures that efficiently represent a family of sparse subsets from a finite ordered universe $U$ \cite{ZDDsandEnumProbs,ZDD}. In the context of graph theory, the diagram may also represent a family of subgraphs $G' \subseteq G$ for a graph $G = (V, E)$. The frontier-based search algorithm provides a way to construct the diagram representing subgraphs, particularly paths, in a graph by setting the universe as the set of edges $U = E$ \cite{FBS}.

While the \texttt{TdZdd} library is commonly used for ZDD construction and set operations, its C++ implementation posed compatibility issues with our Python pipeline. Hence, we adopted the \texttt{graphillion} library instead, which offers a Python interface for ZDD operations. Although the standardized moments were calculated post hoc in our implementation, direct computation using ZDD operations, as demonstrated by \cite{brian2024_zdd_moments}, is also possible.

\subsection{Confounder Detector}
Detecting unmeasured confounders is crucial especially for LiNGAM, which assumes the absence of unmeasured confounders. In this section, we develop a classifier that detects the presence of unmeasured confounders. The features used were the 3rd to the 30th standard moments of the log-transformed path distribution and the target is binary -- 1 if there are confounders and 0 if none. We then train different classifier models as shown in Table \ref{tab:confounder_detector_model_comparison}. Note that for demonstration purposes, these models were not fine-tuned, hence there is room for improving these results. Regardless, the models demonstrate good performance, with CatBoost achieving the highest AUC of $0.78$ that indicates good discrimination between the classes.

\begin{table}
  \caption{Confounder Detectors Trained with Various Classifiers}
  \begin{tabular}{r|rrrrr}
\toprule
        Model &    AUC &  Precision &  Recall &  Accuracy &  Optimal Threshold \\
\midrule
         $k$NN & 0.6537 &     0.5918 &  0.7365 &    0.6142 &             0.4000 \\
Random Forest & 0.7608 &     0.6883 &  0.7465 &    0.7042 &             0.3300 \\
      XGBoost & 0.7665 &     0.6955 &  0.7400 &    0.7080 &             0.3278 \\
     \textbf{CatBoost} & \textbf{0.7829} &     \textbf{0.6957} &  \textbf{0.7740} &    \textbf{0.7178} &             0.3271 \\
     AdaBoost & 0.6997 &     0.6557 &  0.7540 &    0.6790 &             0.4972 \\
\bottomrule
\end{tabular}

  \label{tab:confounder_detector_model_comparison}
\end{table}

\begin{figure}[h]
\centering
\includegraphics[width=\linewidth]{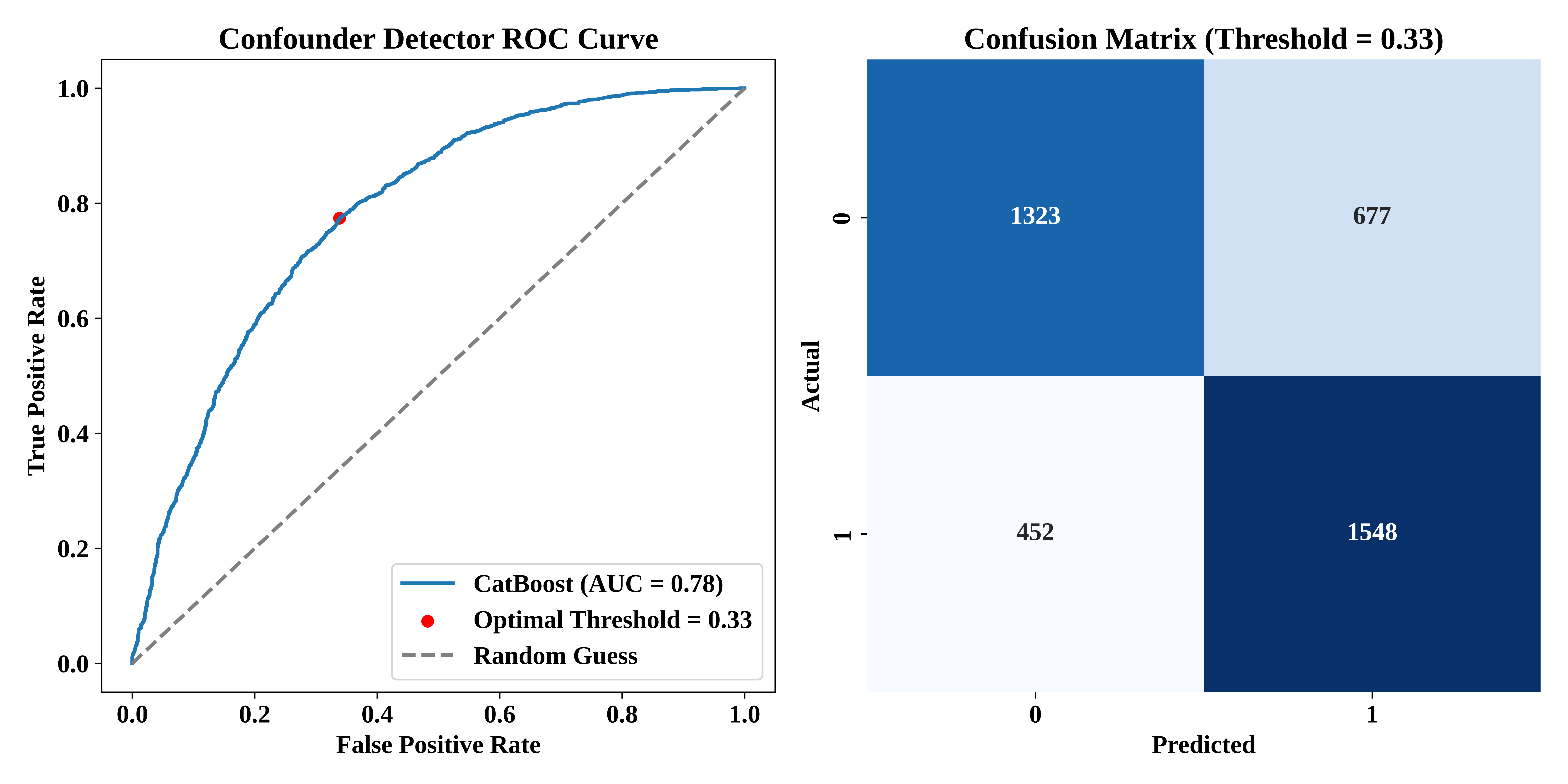}
\caption{ROC Curve and Confusion Matrix of the Confounder Detector (CatBoost)}
\Description{The figure illustrates the Receiver Operating Characteristic (ROC) curve and confusion matrix for the confounder detector model. Employing CatBoost and assessed on the test set, the optimal threshold, identified through Youden's J statistic, maximizes the difference between true positive and false positive rates. The associated confusion matrix is derived from this threshold, showcasing the CatBoost model's remarkable statistics ($0.77$ recall and $0.70$ precision). 
}
\label{fig:confounder_detector_roc_cm}
\end{figure}

Figure \ref{fig:confounder_detector_roc_cm} shows the ROC curve and the confusion matrix from CatBoost evaluated on the test set. The optimal threshold is determined using Youden's J statistic, which selects the threshold that maximizes the difference between the true positive and false positive rates. This threshold can be changed depending on the precision and sensitivity requirements of the use case. For the optimal threshold, we nevertheless see that the CatBoost model exhibits $0.77$ recall and $0.70$ precision.

We have thus shown that it is possible to detect unmeasured confounders with reasonable performance using information from the shape of the path distribution captured through the standardized moments.

\subsection{Sparsity Estimator}
\begin{figure}[h]
\centering
\includegraphics[width=\linewidth]{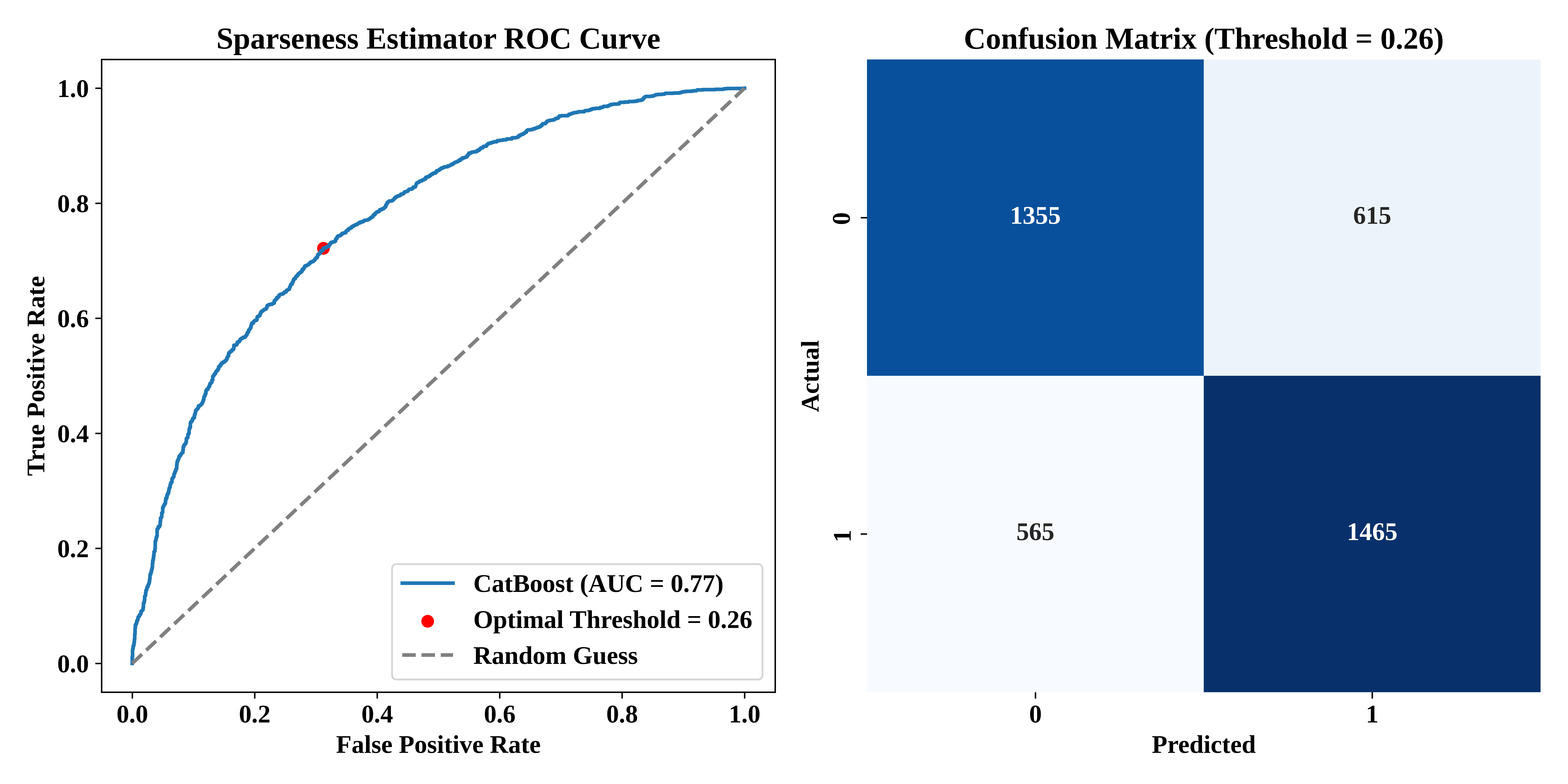}
\caption{ROC Curve and Confusion Matrix of the Sparsity Estimator (CatBoost)}
\Description{The figure illustrates the Receiver Operating Characteristic (ROC) curve and confusion matrix for the sparsity estimator model. Employing CatBoost and assessed on the test set, the optimal threshold, identified through Youden's J statistic, maximizes the difference between true positive and false positive rates. The associated confusion matrix is derived from this threshold, showcasing the CatBoost model's remarkable statistics ($0.72$ recall and $0.70$ precision).}
\label{fig:sparseness_estimator_roc_cm}
\end{figure}

Another task we explore is predicting graph sparsity using standard moments. Using the same features as the confounder detector, we initially create a classifier predicting whether sparsity is greater than $0.5$. Table \ref{tab:sparseness_estimator_model_comparison} presents the results from different models. Similar to the confounder detector, CatBoost achieved the highest AUC of $0.77$. While the precision and recall of CatBoost are remarkable, AdaBoost offers slightly better precision and XGBoost provides slightly better recall. Choosing between them depends on the specific use case. Figure \ref{fig:sparseness_estimator_roc_cm} displays the ROC curve and confusion matrix for CatBoost. Once again, it demonstrates good AUC, precision, and recall, indicating effective discrimination.

\begin{table}
  \caption{Sparsity Estimators Trained with Various Classifiers}
  \begin{tabular}{r|rrrrr}
\toprule
        Model &    AUC &  Precision &  Recall &  Accuracy &  Optimal Threshold \\
\midrule
         $k$NN & 0.6262 &     0.5949 &  0.6054 &    0.5905 &             0.4000 \\
Random Forest & 0.7286 &     0.7174 &  0.5640 &    0.6660 &             0.3400 \\
      XGBoost & 0.7551 &     0.6773 &  \textbf{0.7443} &    0.6902 &             0.2161 \\
     \textbf{CatBoost} & \textbf{0.7744} &     0.7043 &  0.7217 &    \textbf{0.7050} &             0.2647 \\
     AdaBoost & 0.7579 &     \textbf{0.7285} &  0.6438 &    0.6975 &             0.4958 \\
\bottomrule
\end{tabular}

  \label{tab:sparseness_estimator_model_comparison}
\end{table}

In addition to the classifier, we explored creating a regression model using the same features with the target being the actual sparsity ranging from 0 to 1. Table \ref{tab:sparseness_regression_results} compares results from different models. Among them, AdaBoost delivered the best coefficient of determination $R^2$ of $0.16$, indicating that the shape of the distribution may explain some of the variance in sparsity. AdaBoost also yields a root-mean-square error (RMSE) of $0.27$, reasonably estimating sparsity.

\begin{table}
  \caption{Regression Results for Sparsity Estimation}
  \begin{tabular}{r|rrr}
\toprule
        Model &  $R^2$ &   RMSE &    MAE \\
\midrule
         $k$NN &    -0.0804 & 0.3002 & 0.2460 \\
Random Forest &     0.0824 & 0.2766 & 0.2240 \\
      XGBoost &     0.0645 & 0.2793 & 0.2269 \\
     CatBoost &     0.1027 & 0.2736 & \textbf{0.2229} \\
     \textbf{AdaBoost} &     \textbf{0.1568} & \textbf{0.2652} & 0.2252 \\
\bottomrule
\end{tabular}
  
  \label{tab:sparseness_regression_results}
\end{table}

\subsection{Performance Predictors}
So far, we have created models for detecting confounders and graph sparsity. These two properties are crucial for evaluating the performance of DirectLiNGAM and LiNGAM-SPP, as cases involving confounders or sparse graphs tend to pose greater challenges and lead to the poor performance of these methods. Further, we want to test if it is possible to directly predict the performance of these methods. In line with this, we develop classifiers to predict whether PLR DirectLiNGAM and PLR LiNGAM-SPP would produce the correct causal ordering. In other words, the target is 1 if the predicted causal order is entirely correct. Table \ref{tab:performance_predictor_classification} showcases the results.

\begin{table}
  \caption{Performance Predictors Trained with Various Classifiers}
  \begin{tabular}{r|rrrr|rrrr}
\toprule
\multicolumn{1}{c}{} & \multicolumn{4}{c}{LiNGAM-SPP} & \multicolumn{4}{c}{DirectLiNGAM} \\
\hline
Model &        AUC & Precision &  Recall & Accuracy &     AUC & Precision &  Recall & Accuracy \\
\midrule
$k$NN          &     0.641 &    0.155 &  0.613 &   0.617 &            0.638 &    0.147 &  0.594 &   0.615 \\
Random Forest &     0.822 &    0.272 &  0.752 &   0.768 &            \textbf{0.821} &    0.239 &  \textbf{0.801} &   0.725 \\
XGBoost       &     0.799 &    0.237 &  \textbf{0.774} &   0.721 &            0.800 &    \textbf{0.311} &  0.648 &   \textbf{0.821} \\
CatBoost      &     \textbf{0.822} &    \textbf{0.333} &  0.698 &   \textbf{0.826} &            \textbf{0.821} &    0.306 &  0.711 &   0.809 \\
AdaBoost      &     0.790 &    0.292 &  0.672 &   0.799 &            0.789 &    0.269 &  0.711 &   0.777 \\
\bottomrule
\end{tabular}

  \label{tab:performance_predictor_classification}
\end{table}

For the LiNGAM-SPP performance predictor, CatBoost delivered the best AUC of 0.82. Given the relatively small number of instances with exact correct ordering, it is more likely to get some of the ordering wrong than to get everything right. Recall is therefore reasonable at $0.70$ but precision is not very high at $0.33$, indicating a significant number of false positives.

Meanwhile, for the DirectLiNGAM performance predictor, both the Random Forest and CatBoost provided similar AUC metrics of $0.82$. The Random Forest has the best recall at $0.80$ while another model, XGBoost, exhibits the highest precision at $0.31$. Similar to the LiNGAM-SPP case, the sensitivity of the models is good but precision is not as high, suggesting a notable presence of false positives. 

We also explored fitting regression models to the ordering score of LiNGAM-SPP and DirectLiNGAM, with the results shown in Table \ref{tab:performance_predictor_regression}. For the LiNGAM-SPP case, AdaBoost emerged as the best model, achieving a $0.21$ $R^2$ that indicates its ability to explain some variance in the ordering score. AdaBoost also boasted the lowest RMSE of $0.18$, reasonably pinpointing the ordering score within a reasonable range. For DirectLiNGAM, AdaBoost also outperformed other models, achieving an $R^2$ of $0.21$ and an RMSE of $0.18$.

\begin{table}
  \caption{Regression Results for Performance Prediction}
  \begin{tabular}{r|rrr|rrr}
\hline
\multicolumn{1}{c}{} & \multicolumn{3}{c}{LiNGAM-SPP} & \multicolumn{3}{c}{DirectLiNGAM} \\
\hline
Model &  $R^2$ &    RMSE &     MAE &    $R^2$ &    RMSE &     MAE \\
\midrule
$k$NN          &    -0.1012 &  0.2150 &  0.1674 &      -0.0853 &  0.2149 &  0.1672 \\
Random Forest &     0.0836 &  0.1961 &  0.1472 &       0.0853 &  0.1973 &  0.1487 \\
XGBoost       &     0.0388 &  0.2008 &  0.1499 &       0.0319 &  0.2030 &  0.1521 \\
CatBoost      &     0.1162 &  0.1926 &  0.1441 &       0.1009 &  0.1956 &  0.1465 \\
AdaBoost      &     \textbf{0.2109} &  \textbf{0.1820} &  \textbf{0.1436} &       \textbf{0.2096} &  \textbf{0.1834} &  \textbf{0.1449} \\
\bottomrule
\end{tabular}

  \label{tab:performance_predictor_regression}
\end{table}

\subsection{Remarks on Path Sampling}
One drawback of path enumeration is the factorial increase of the number of edges with the number of features. To address the impracticality of using the entire path distribution especially in cases with large $p$, we opt for a more practical solution—sampling from the path distribution to compute the standardized moments. As mentioned earlier, ZDDs provide this functionality by sampling from the subset of graphs it represents. Here, we demonstrate the trade-off between path sample size and model performance, revealing that sufficiently good performance for the models discussed earlier may still be achieved without resorting to the entire path distribution.

Table \ref{tab:path_sampling_auc} presents the performance of the models under various path sampling sizes. In this demonstration, we limit the training and testing to $p=7$ since $7! = 5040$, which is sufficiently large for our evaluation. The table illustrates that the AUC of the model generally increases with the path sample size, approaching the performance of the full path distribution. For instance, even at a path sample size of 2500, the confounder detector and sparsity estimator already achieved AUCs that are sufficiently close to those obtained using the full path distribution. While the performance predictors for the same path sample size still have a slightly larger gap, the performance is acceptable. For larger values of $p$, even a fraction of the possible paths may already yield satisfactory performance.

\begin{table}
  \caption{Model Performance Under Different Path Sampling Sizes}
  \begin{tabular}{rcccc}
\toprule
 Path Sample&  Confounder &  Sparseness &  \multicolumn{2}{c}{Performance Predictors} \\
 Size &  Detector &  Estimator &  LiNGAM-SPP &  DirectLiNGAM \\
\midrule
         100 &           0.7453 &     0.7317 &                 0.6811 &                   0.6880 \\
         250 &           0.7498 &     0.7553 &                 0.7436 &                   0.7241 \\
         500 &           0.7824 &     0.7734 &                 0.7772 &                   0.7792 \\
        1000 &           0.7979 &     0.7856 &                 0.7980 &                   0.7969 \\
        2500 &           0.8183 &     0.7932 &                 0.8140 &                   0.8075 \\
\hline
        All Paths (5040) &           0.8204 &     0.7939 &                 0.8480 &                   0.8538 \\
\bottomrule
\end{tabular}

  \label{tab:path_sampling_auc}
\end{table}

\section{Conclusion}
We address a critical limitation of LiNGAM-SPP by eliminating the need for parameter tuning. Our modification, which uses the PLR \cite{hyvarinen13a} in place of the $k$NN-based mutual information estimators from earlier LiNGAM-SPP studies \cite{suzuki_lingam_mmi, suzuki2024generalization}, not only eliminated the need for fine-tuning but also demonstrated superior performance and computational efficiency on simulated data. Furthermore, the modified LiNGAM-SPP we propose, PLR LiNGAM-SPP, proves to be more stable than DirectLiNGAM, significantly outperforming it in simulations with more features. Our method performs at least as well as DirectLiNGAM on real-world data and in some cases, outperforms it by capturing more required edges and fewer forbidden edges.

We also introduce a functionality to LiNGAM-SPP for incorporating prior knowledge. Our approach allows for specifying relative orderings rather than exact edges, offering increased flexibility. The integration of prior knowledge improved performance and reduced the number of edges traversed.

Furthermore, we extend the LiNGAM-SPP method by leveraging the entire path distribution. The standardized moments of the path distribution were used as features to create models that estimate certain properties of the true causal graph. We created a confounder detector, a sparsity estimator, and predictors for DirectLiNGAM and PLR LiNGAM-SPP performance. The confounder detector and sparsity estimator showed robust performance, achieving high AUCs of $0.78$ and $0.77$, respectively. Although the DirectLiNGAM and LiNGAM-SPP performance predictors achieved AUCs of $0.822$ and $0.821$, respectively, they exhibited a higher rate of false positives due to the inherent challenge in getting the exact causal ordering.

We acknowledge the limitations of our study, such as the intractability of path enumeration for large $p$, which led us to limit our demonstration to a small number of features. This limitation is somewhat addressed using the sampling feature of ZDDs, but constructing the ZDD itself becomes more difficult and requires more memory as $p$ increases.

To close, our study proposes enhancements to the current LiNGAM-SPP and demonstrates other ways to leverage the shortest path problem formulation of causal discovery frameworks. While further research is needed to address the limitations and provide theoretical grounding, our results offer a promising direction for inferring notable properties of the true causal graph from observational data. 

\bibliographystyle{ACM-Reference-Format}
\bibliography{refs}

\end{document}